# Evaluating Ukrainian Grammatical Competence in Language Models: Probability Versus Prompting

Dmytro Babarytskyi[1,*], Oleksandr Babarytskyi[2] and Vadym Savanevych[1]

*1 Kharkiv National University of Radio Electronics, Nauky Ave. 14, 61166 Kharkiv, Ukraine*

*2 Military Law Institute, Yaroslav Mudryi National Law University, Dynamivska St. 4, 61023 Kharkiv, Ukraine*

* Corresponding author.
dmytro.babarytskyi@nure.ua (D. Babarytskyi); aleksandr.babaritskiy@ukr.net (O. Babarytskyi);
vadym.savanevych1@nure.ua (V. Savanevych)
0009-0009-3614-9601 (D. Babarytskyi); 0000-0002-1690-1837 (O. Babarytskyi); 0000-0001-8840-8278 (V. Savanevych)

**Abstract**

Organisations deploying large language models in Ukrainian-language information systems need to know whether a model produces normative Ukrainian. This paper establishes a protocol for measuring that reliably and applies it to 19 models. The instrument is a released set of 1,444 Ukrainian minimal pairs covering the vocative case, numeral–noun and adjective agreement, verb government and calques from russian, with subject–verb agreement as a control; a quarter of the set was checked against the answer key in two independent native-speaker ratings, which agreed at Cohen's $\kappa$ = 0.963. Every pair is scored twice: by comparing the probabilities a model assigns to the two sentences, and by asking the model directly, with the sentences presented in both orders so that answer-position effects are measured instead of being absorbed into the score. Three results follow. Order consistency — the share of pairs answered identically when the sentences are exchanged — falls below 0.05 for ten of the nineteen models, and what separates the usable models is instruction tuning, not scale. The vocative case is systematically inverted: 0.153 accuracy against 0.951 on the control, with models preferring the non-normative nominative form of address. Ukrainian adaptation substantially reduces this deficit, measured against the exact checkpoints two adapted models were trained from.

## 1. Introduction

Large language models are increasingly embedded in information systems that serve Ukrainian-speaking users: document processing, customer-facing assistants, automated correspondence, content generation. For such deployments the relevant question is not whether a model can produce Ukrainian text, which all current models can, but whether the Ukrainian it produces conforms to the standard norm. An assistant that writes «на протязі місяця» instead of «протягом місяця», or addresses a user as «Олег» where «Олеже» is required, is fluent and wrong in a way that a fluency metric will not detect.

Answering that question requires an evaluation method, and the method most commonly used is to prompt the model: present several variants, ask which is correct, count the correct answers. This is convenient, it works with closed models available only through an interface, and it mirrors how humans are tested. It also carries an assumption that is rarely checked — that the model's answer depends on the content of the options and not on their arrangement.

We test that assumption for Ukrainian and find that it fails for a large class of models, and that what governs the failure is not model size but whether the model has been instruction-tuned. Presenting the same pair of sentences twice with the positions exchanged, ten of the nineteen models we evaluate give the same answer on more than 95% of pairs regardless of which sentence occupies which position — among them a 12B model, while a 4B instruction-tuned model answers by content two times out of three. For these models a prompted accuracy score measures the strength of a positional preference. One model in our study reaches the third-highest prompted accuracy in the entire evaluation while answering consistently on fewer than one pair in five.

The alternative is to compare the probabilities a model assigns to the two sentences, which requires access to the model's outputs but does not depend on its willingness or ability to follow an instruction. To apply this method at scale one needs minimal pairs: sentences differing in exactly one element, one normative and one not. For Ukrainian, the only existing resource of this kind covers subject–verb agreement alone, as part of a multilingual effort, and that is the one phenomenon current models already handle almost perfectly.

This paper contributes the following.

**A protocol for measuring Ukrainian normative usage.** Probability scoring over minimal pairs, paired with an order-consistency diagnostic that establishes when a prompted score carries information about grammar. The diagnostic costs one extra presentation per item and tells a practitioner which evaluation channel their model supports: for base checkpoints at any scale, and for instruction-tuned models below roughly 4B parameters, the prompted channel reports answer position, and probability scoring is the one that measures the norm.

**A diagnosis of what current models get wrong.** The vocative case, a productive and obligatory feature of standard Ukrainian, is systematically inverted in multilingual models: they assign higher probability to the non-normative nominative form of address in approximately six cases out of seven, while scoring 0.951 on the control phenomenon. Length normalisation reduces the effect without removing it. Calque failures are concentrated in a small set of entrenched items whose non-normative variants dominate everyday usage.

**A quantified remedy.** Ukrainian adaptation substantially reduces these specific deficits, measured against the exact checkpoints two adapted models were trained from, on identical items under identical configurations. The gain is concentrated: the control phenomenon does not move at all, and almost the entire improvement falls on the vocative and on calques.

The pair set, the scoring harness, the analysis code and the raw per-pair scores for all 19 models are publicly released, so that every number reported here can be recomputed and the instrument can be applied to models we did not evaluate.

## 2. Related Works

### 2.1. Minimal-Pair Benchmarks

Minimal pairs have become a standard instrument for probing the grammatical knowledge of language models. A minimal pair consists of two sentences that differ in exactly one element, one of which violates a grammatical rule; a model is credited with knowing the rule when it assigns the higher probability to the well-formed member. BLiMP [1] established the format for English with 67 paradigms, and the design has since been carried to other languages: QFrBLiMP [2] applies it to Quebec French, confirming that the format transfers to varieties whose norms differ from a dominant standard.

For Ukrainian, coverage remains thin. MultiBLiMP 1.0 [3] spans 101 languages and includes Ukrainian, but its pipeline is built on Universal Dependencies and UniMorph and is therefore restricted to subject–verb agreement. That is a single paradigm out of the many where Ukrainian norms are contested in practice, and it is precisely the paradigm that current models already handle well. We include subject–verb agreement in our set as a **control** and not as an object of study, so that our measurements can be checked against an existing resource.

### 2.2. Evaluation Resources for Ukrainian

The resource closest to our work is the expert-curated benchmark of Galeshchuk et al. [4], prepared by professional linguists and by the National Commission for State Language Standards. It contains 347 multiple-choice questions on grammar, lexical norms and orthography, and evaluates Ukrainian-adapted, multilingual and large-scale models under zero-shot and few-shot prompting. Their findings are sobering: smaller models reach at most 42.1% accuracy and the strongest large model 59.6%.

Our work complements theirs and differs in the unit of measurement. Their items ask which of four or five rows contains only correct forms, so a single item bundles up to a dozen individual judgements into one binary outcome. This is appropriate for assessing overall proficiency, but it cannot localise *which* form a model fails on, and with four or five options the chance baseline is around 22%, which compresses the range in which models can be distinguished. A minimal pair isolates exactly one alternation, has a chance baseline of 50%, and permits a per-phenomenon and per-item breakdown. The two instruments answer different questions about the same competence.

Other Ukrainian benchmarks target other abilities. ZNO-Eval [5] measures reasoning using standardised examination tasks, UAlign [6] evaluates ethical alignment, and UA-Legal-Bench [7] covers legal reasoning. Karpo and Chernodub [8] study prompting for minimal-edit grammatical error correction, which addresses the same phenomena from the correction side rather than the evaluation side. None of these targets normative usage with per-phenomenon resolution, and none covers calques from russian — a class of forms that is fluent, frequent in everyday usage, and non-normative, and which by construction cannot appear in an English or russian benchmark.

### 2.3. Probability Measurements and Prompting

A model can be asked about grammar in two ways: by comparing the probabilities it assigns to the two sentences, or by asking it directly which sentence is correct. Hu and Levy [9] showed for English that these two channels dissociate, and that metalinguistic prompting systematically underestimates the knowledge that probability measurements reveal. Ide et al. [10] extend the analysis and report that how the probability is computed — in particular whether it is normalised — materially changes the resulting acceptability judgement.

A complication specific to the prompted channel is position bias. Zheng et al. [11] demonstrate that language models are not robust multiple-choice selectors and carry a systematic preference for particular option positions, and Tang et al. [12] document the same effect in question generation. The practical consequence for grammatical evaluation has not, to our knowledge, been quantified for Ukrainian: if a model answers by position and not by content, a prompted accuracy figure reports the strength of that preference, not knowledge of the norm. We therefore present every pair in both orders and report order consistency alongside accuracy.

### 2.4. Ukrainian-Adapted Models

Two families of Ukrainian-adapted models are publicly available: MamayLM [13], continued pre-training and instruction tuning of Gemma 3 4B by INSAIT, and Lapa LLM [14], a continued pre-train of Gemma 3 12B. Both are evaluated in [4] alongside multilingual baselines. What has not been reported is a **paired** comparison against the exact checkpoint each was adapted from, on identical items, under an identical loading configuration. Without such a comparison, a difference between an adapted model and a multilingual one of another family confounds the effect of adaptation with the effects of family, size and training data. Supplying that comparison, for two independent base–adapted pairs, is one of the contributions of this paper.

## 3. Proposed Methodology

### 3.1. Pair Construction

The benchmark consists of 1,448 generated minimal pairs covering five phenomena of standard Ukrainian plus one control. Each pair is a sentence in which exactly one edited span distinguishes the normative member from the non-normative one: one word, one inflected form, or one construction replaced by another. In 44 pairs the two members contain a different number of words, because the alternating construction has a different length on either side («протягом року» versus «на протязі року»); the released data records the number of edited spans and the word-count equality for every pair, so that any subset can be filtered by these properties.

Three phenomena are generated from templates combined with the pymorphy3 morphological analyser: numeral–noun agreement, adjective–noun agreement, and the subject–verb agreement control. For these, sentence frames are filled with inflected forms produced by the analyser, and the non-normative member is obtained by substituting a form with incorrect number, gender or case. Two phenomena draw on curated lists: the vocative, built from a list of Ukrainian given names with their vocative forms embedded in eight address frames, and verb government, built from a list of verbs whose case requirements differ from those of the corresponding russian verbs. Calques are taken entirely from a curated list, each item instantiated in four contexts.

Generation is deterministic: a fixed random seed selects which instantiations enter the released set, so the procedure reproduces the identical file from the same inputs.

Subject–verb agreement serves as a control, not as an object of study. It is the one Ukrainian phenomenon already covered by an existing multilingual resource [3], so a harness that measures correctly should place models near ceiling on it. Any result on the remaining phenomena is interpreted against that reference point rather than in absolute terms.

### 3.2. Normativity and Exclusions

Whether a form counts as non-normative is decided by published reference works and not by the authors' judgement. For the vocative, numeral–noun and adjective agreement and the subject–verb control, the authority is the Ukrainian Orthography of 2019 [15], which regulates these forms directly. The calque and verb-government lists, which fall outside what an orthography covers, were compiled against Ukrainian usage handbooks: Ponomariv's *Kultura slova* [16], Antonenko-Davydovych's *Yak my hovorymo* [17] and *Antysurzhyk* [18]. Items on which these sources disagree, and items where the alternation marks a difference of nuance and not of norm, were excluded during construction instead of being adjudicated. This is a deliberate narrowing: the benchmark measures agreement with a codified standard, not the full space of acceptable Ukrainian.

### 3.3. Human Validation

A stratified sample of 368 pairs — 25% of the set, including every calque and every verb-government item — was rated independently by the first two authors, both native speakers of Ukrainian. Neither had access to the answer key, and neither used dictionaries during rating. For each pair the rater marked which member was normative, or marked the pair as one where both members are acceptable, and rated how noticeable the difference was on a three-point scale. The completed sheets are released with the data.

**Table 1**
A 368-pair sample, 25% of the set, checked against the answer key in two independent ratings. Agreement is between the two ratings; defect is the share of pairs both placed against the key. Cohen's kappa = 0.963 overall.

| Phenomenon | Rated | Agreement | Defect |
|---|---|---|---|
| Subject–verb (control) | 30 | 1.000 | 0.000 |
| Adjective agreement | 30 | 1.000 | 0.000 |
| Numeral + noun | 30 | 1.000 | 0.000 |
| Verb government | 60 | 0.917 | 0.000 |
| Calques | 188 | 0.995 | 0.021 |
| Vocative | 30 | 0.967 | 0.000 |

Table 1 reports agreement and defect rates per phenomenon. Inter-rater agreement was Cohen's $\kappa = 0.963$ with 98.1% raw agreement. Measured against the key, rater 1 agreed on 97.0% of pairs and rater 2 on 98.9%. A pair was counted as defective only when **both** raters disagreed with the key, since a single disagreement is rater noise while unanimous disagreement indicates a bad item. By that criterion the defect rate is 1.1%.

All four defective pairs belonged to a single group, the alternation «ймовірно» / «вірогідно», which both raters judged to be a difference of nuance and not of normativity. Applying the rule fixed in advance — when both raters reject the key, the whole group is removed, not only the sampled items — that group was dropped in full, leaving **1,444 pairs** for evaluation.

### 3.4. Probability Scoring

The probability score asks whether a model *uses* the norm. A sentence $s$ is tokenised into $t_1 \ldots t_n$ and scored by the total log-probability its tokens receive under the model:

$$\log P(s) = \Sigma \log p(t_i \mid t_1 \ldots t_{i-1}) \quad (1)$$

A pair consisting of the normative sentence $s^+$ and its non-normative counterpart $s^-$ counts as correct when the normative member takes the higher total, where $I[\cdot]$ is the indicator function:

$$y = I[\ \log P(s^+) > \log P(s^-)\ ] \quad (2)$$

Several phenomena change the number of tokens in the sentence — vocative forms are on average 0.46 tokens longer than the corresponding nominatives — so we also report a variant in which each total is divided by its own token count $n^+$ or $n^-$ before the comparison:

$$y' = I[\ \log P(s^+) / n^+ > \log P(s^-) / n^-\ ] \quad (3)$$

Both variants are reported throughout; where they diverge, the divergence is itself informative.

### 3.5. Prompted Scoring and Order Consistency

The prompted score asks whether a model *knows* the norm. The model is shown both sentences, labelled А and Б, and asked in Ukrainian which conforms to the norms of standard Ukrainian; we then compare the log-probabilities it assigns to answering "А" and to answering "Б". For instruction-tuned models the question is wrapped in the model's own chat template.

Writing $q$ for the question as presented and $\ell^+$, $\ell^-$ for the letters that label the normative and the non-normative sentence in that presentation, the margin in favour of the norm is

$$m(q) = \log P(\ell^+ \mid q) - \log P(\ell^- \mid q) \quad (4)$$

Every pair is asked **twice, in both orders**, giving two presentations $q_1$ and $q_2$ of the same pair. The prompted answer counts as correct when the mean margin favours the normative sentence:

$$z = I[\ (\ m(q_1) + m(q_2)\ ) / 2 > 0\ ] \quad (5)$$

The two presentations also measure something the mean hides. Over the $N$ pairs of a phenomenon, *order consistency* is the share on which the model chooses the same sentence in both presentations:

$$C = (1 / N)\ \Sigma\ I[\ \text{sgn}\ m(q_1) = \text{sgn}\ m(q_2)\ ] \quad (6)$$

Consistency is not a secondary diagnostic but a precondition for interpreting the prompted score at all. A model that answers by position rather than by content will produce a prompted accuracy determined by the strength of its positional preference, not by its grammatical knowledge. We therefore report consistency for every model, restrict the comparison between the two scoring methods to models whose consistency exceeds 0.6, and additionally report accuracy computed only on the order-consistent subset.

### 3.6. Models and Configuration

We evaluate 19 publicly available models: six from the Gemma family [19], five from Qwen2.5 [20] and five from Llama 3 [21], and three Ukrainian-adapted models — MamayLM [13] (Gemma 3 4B) and two Lapa checkpoints [14] (Gemma 3 12B, base and instruction-tuned). Where both a base and an instruction-tuned variant of the same checkpoint exist, both are evaluated, giving seven base and nine instruction-tuned models among the multilingual ones.

Two pairs are of particular interest, because in each the adapted model and its own starting checkpoint are both evaluated: lapa-12b-pt against gemma-3-12b-pt, and MamayLM against gemma-3-4b-it. Within each pair the two models are loaded in an identical configuration, so the comparison isolates the effect of Ukrainian adaptation from the effects of family, size and precision.

All experiments run on a single 16 GB T4 GPU, with two T4s used for the 12B checkpoints. Models above 3B parameters are loaded in 4-bit NF4 quantisation. Gemma 3 is trained in bfloat16, which the T4 does not support; at 12B its activations overflow float16 and every log-probability returns NaN, so the 12B checkpoints are loaded and computed in float32. The image encoder of the multimodal Gemma 3 checkpoints is released after loading, as it plays no part in scoring text. Precision settings are recorded per model and are identical within each controlled pair.

### 3.7. Statistical Treatment

Accuracies are reported with 95% Wilson score intervals. Comparisons between two measurements on the same pairs — the two scoring methods for one model, or one model against another — use McNemar's test with continuity correction. With $b$ the number of pairs the first measurement gets right and the second wrong, and $c$ the number for which the reverse holds, the statistic is

$$\chi^2 = ( | b - c | - 1 )^2 / ( b + c ) \tag{7}$$

evaluated against the chi-square distribution with one degree of freedom. Only the discordant pairs enter it, which is what makes it the appropriate test for paired binary outcomes: the pairs both measurements agree on carry no information about a difference between them.

### 3.8. Availability

The pair set, the scoring harness and the analysis scripts are released at github.com/DmytroBabarytskyi/ukrainian-minimal-pairs, with the data under CC BY 4.0 and the code under the MIT licence. The raw per-pair scores for all 19 models are included, so every figure and table in this paper can be recomputed without access to a GPU.

## 4. Results and Discussion

### 4.1. Accuracy Across Phenomena

Table 2 and Figure 1 report accuracy for each phenomenon averaged over all 19 models. The control phenomenon, subject–verb agreement, reaches 0.951 under probability scoring, confirming that the harness measures what it is meant to measure and placing our results on the same footing as MultiBLiMP. Adjective agreement (0.869) and numeral–noun agreement (0.845) follow, then verb government (0.768) and calques (0.773).

**Table 2**
Accuracy per phenomenon, averaged over all 19 models. Prob. is the sum of token log-probabilities, Prob./tok the length-normalised variant.

| Phenomenon | Pairs | Prob. | Prob./tok | Prompt |
|---|---|---|---|---|
| Subject–verb (control) | 300 | 0.951 | 0.886 | 0.838 |
| Adjective agreement | 300 | 0.869 | 0.822 | 0.785 |
| Numeral + noun | 300 | 0.845 | 0.839 | 0.756 |
| Verb government | 60 | 0.768 | 0.664 | 0.648 |
| Calques | 184 | 0.773 | 0.649 | 0.681 |
| Vocative | 300 | 0.153 | 0.323 | 0.325 |

The vocative breaks the pattern completely, at **0.153** (Figure 2). This is not a low score on a difficult phenomenon; it is a systematic inversion. A model choosing at random would score 0.5. Models assign higher probability to the non-normative nominative form of address — «Олег, зайди» over «Олеже, зайди» — in roughly six cases out of seven. Restricted to the sixteen multilingual models, the figure is 0.079, with the best of them at 0.267 and the worst at 0.013.

Token count is a partial explanation but not the explanation. Vocative forms are on average 0.46 tokens longer than the corresponding nominatives, and length-normalised scoring raises the vocative to 0.323 overall and 0.227 for the multilingual models. The correction is substantial and still leaves the result far below chance. The decisive evidence is lapa-12b-pt, a Ukrainian-adapted model: under length-normalised scoring it reaches **0.950** on the vocative. A measurement artefact would depress every model equally; instead, the models trained on Ukrainian text handle the phenomenon almost perfectly, and the multilingual ones systematically prefer the wrong form.

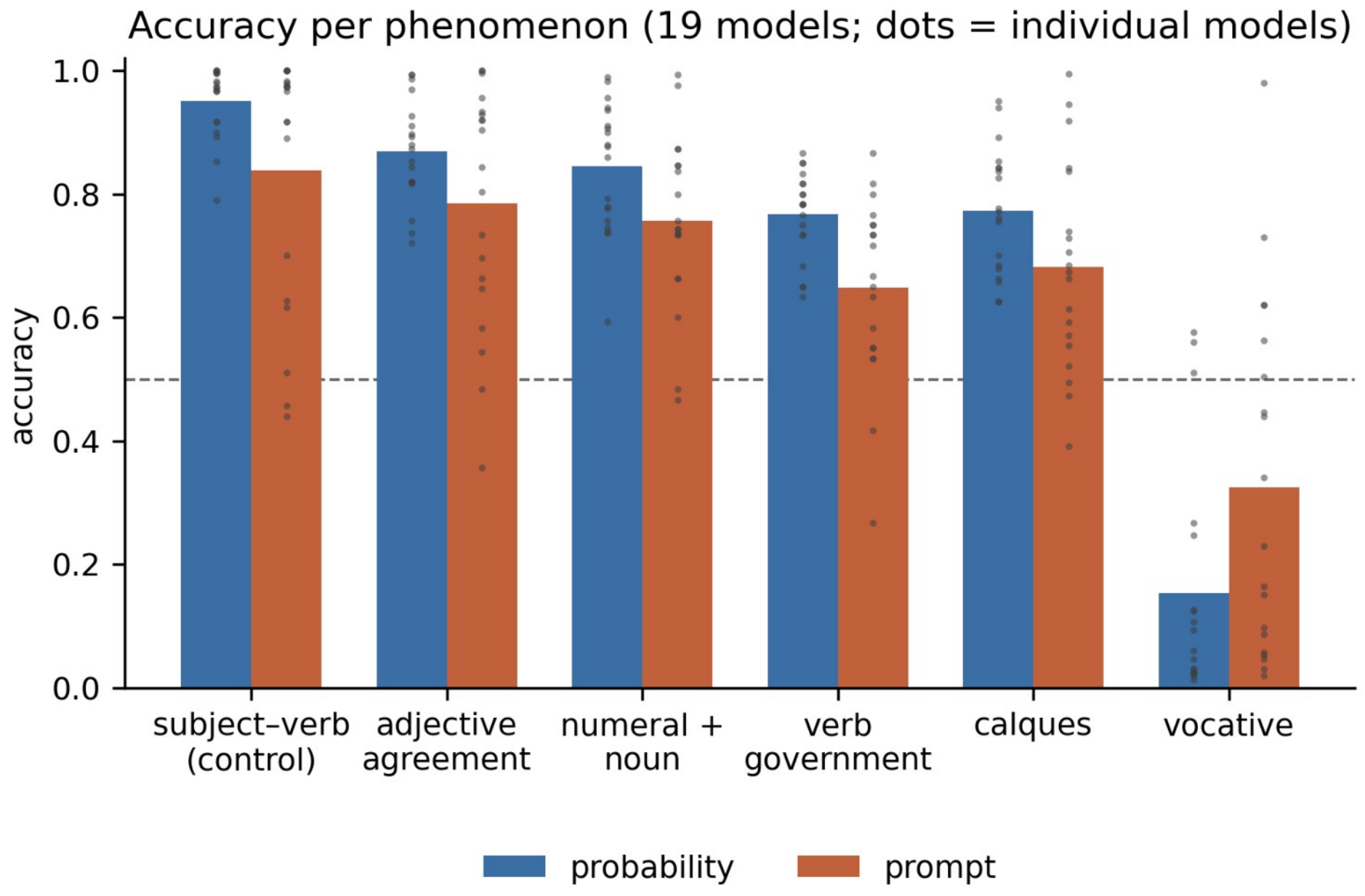

**Figure 1:** Accuracy per phenomenon under both scoring methods, averaged over the 19 models. Dots are individual models.

Feminine names are slightly harder than masculine (0.136 against 0.164), which is expected: for many feminine names the vocative is a less frequent alternation in web text than for masculine ones.

## 4.2. The Prompted Channel Largely Measures Answer Position

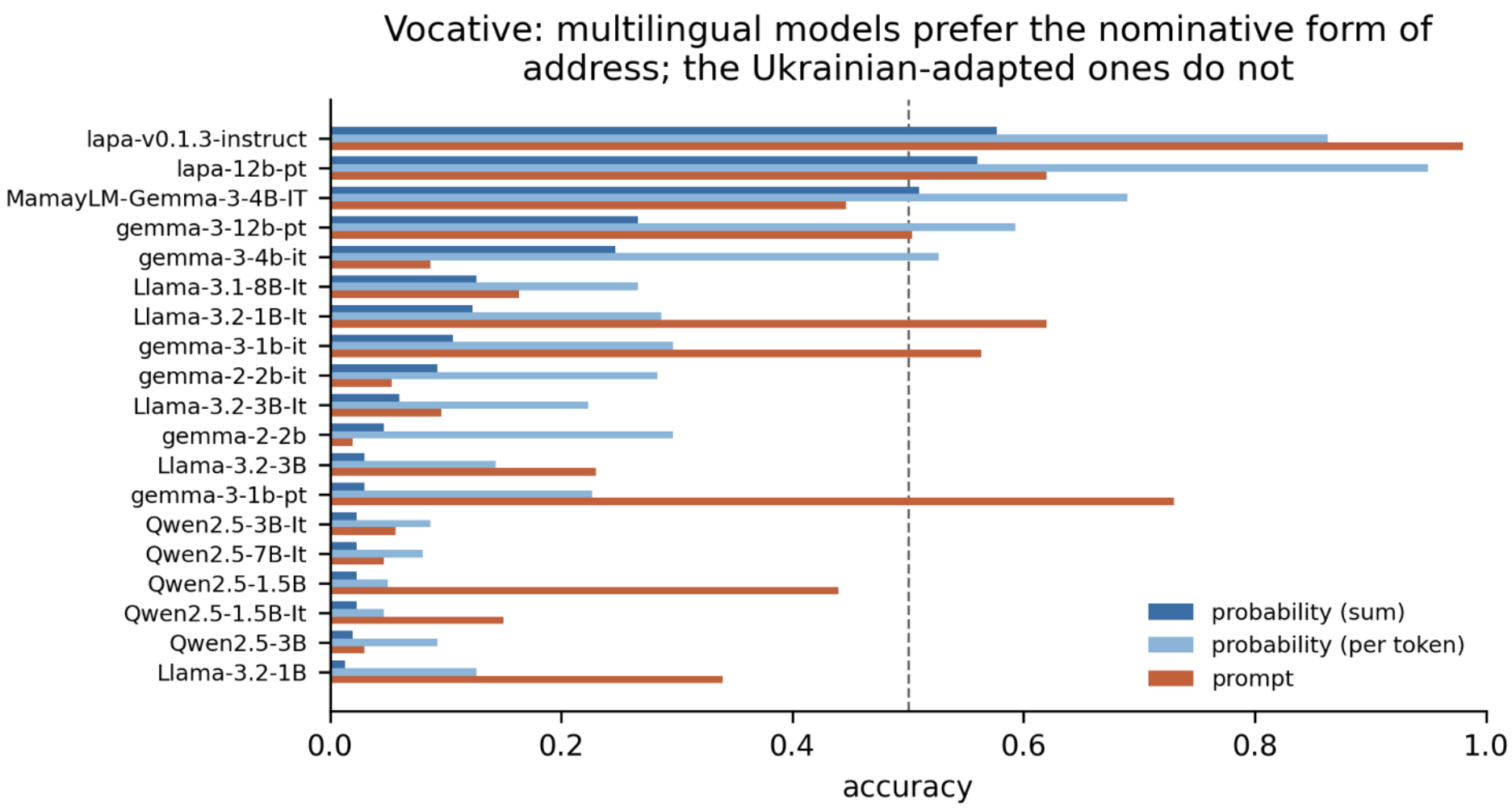

**Figure 2:** Vocative accuracy per model under the three scores. Only the Ukrainian-adapted models approach chance level.

Every pair was presented twice, with the two sentences exchanged. Order consistency — the share of pairs for which a model selects the same sentence in both presentations — is reported per model in Table 3 and Figure 3. It averages **0.242** across the 19 models, and falls below 0.05 for **ten of them**. For those ten, the model names the same letter regardless of which sentence it labels.

Consequently, a prompted accuracy figure for such a model does not report grammatical knowledge. google/gemma-3-12b-pt is the clearest illustration: its prompted accuracy is 0.846, the third highest in the study, while its order consistency is 0.192. On the 19% of pairs it answers consistently, its accuracy is exactly 1.000. The high average is arithmetic: a strong positional preference, combined with the distribution of margins, produces a number that looks like competence.

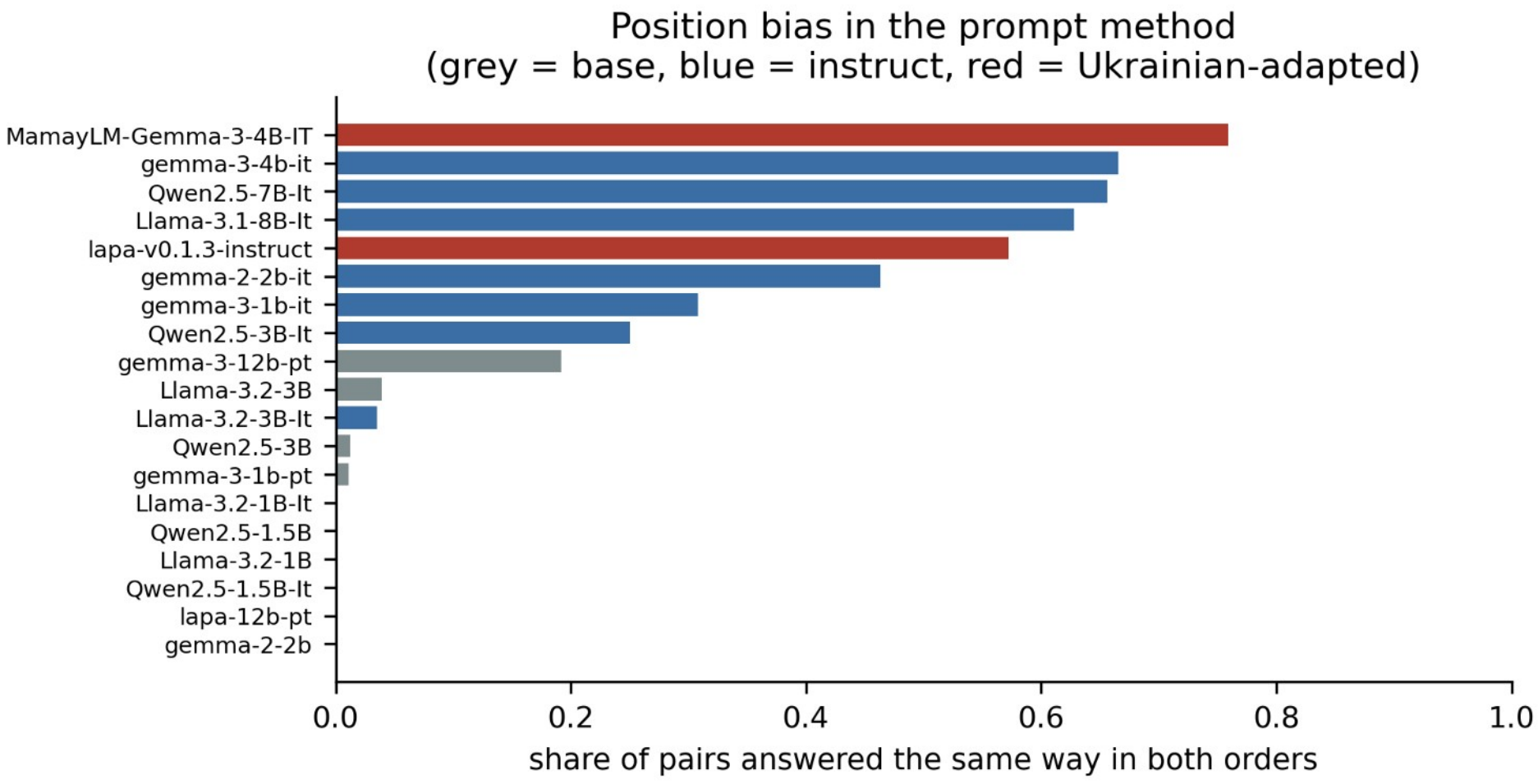

**Figure 3:** Order consistency per model: the share of pairs answered the same way in both presentation orders.

Across all models, accuracy computed only on order-consistent pairs is 0.810, against 0.676 for the unrestricted prompted score.

**Table 3**
Accuracy over 1,444 minimal pairs. PT = base, IT = instruction-tuned; (ua) marks Ukrainian-adapted models. Order consistency is the share of pairs answered the same way when the two sentences are swapped.

| Model | Type | Prob. | Prompt | Order cons. |
|---|---|---|---|---|
| gemma-3-1b-pt | PT | 0.717 | 0.475 | 0.01 |
| gemma-3-1b-it | IT | 0.698 | 0.659 | 0.31 |
| gemma-2-2b | PT | 0.740 | 0.604 | 0.00 |
| gemma-2-2b-it | IT | 0.722 | 0.658 | 0.46 |
| gemma-3-4b-it | IT | 0.766 | 0.731 | 0.67 |
| gemma-3-12b-pt | PT | 0.801 | 0.846 | 0.19 |
| lapa-12b-pt (ua) | PT | 0.889 | 0.758 | 0.00 |
| lapa-v0.1.3-instruct (ua) | IT | 0.886 | 0.986 | 0.57 |
| Llama-3.2-1B | PT | 0.684 | 0.528 | 0.00 |
| Llama-3.2-1B-Instruct | IT | 0.578 | 0.526 | 0.00 |
| Llama-3.2-3B | PT | 0.717 | 0.683 | 0.04 |
| Llama-3.2-3B-Instruct | IT | 0.640 | 0.605 | 0.04 |
| Llama-3.1-8B-Instruct | IT | 0.736 | 0.738 | 0.63 |
| MamayLM-Gemma-3-4B-IT (ua) | IT | 0.868 | 0.867 | 0.76 |
| Qwen2.5-1.5B | PT | 0.630 | 0.519 | 0.00 |
| Qwen2.5-1.5B-Instruct | IT | 0.612 | 0.609 | 0.00 |
| Qwen2.5-3B | PT | 0.628 | 0.681 | 0.01 |
| Qwen2.5-3B-Instruct | IT | 0.640 | 0.695 | 0.25 |
| Qwen2.5-7B-Instruct | IT | 0.651 | 0.670 | 0.66 |

What separates the usable models from the rest is instruction tuning, not size. No base checkpoint reaches 0.2 consistency at any scale: lapa-12b-pt, a 12B model, answers consistently on 0.0% of pairs, and gemma-3-12b-pt on 19.2%. All four models above the 0.6 threshold are instruction-tuned, and among instruction-tuned models consistency grows with scale, from 0.308 at 1B to 0.666 at 4B. What instruction tuning most reliably adds here is the ability to answer the question that was asked, rather than knowledge of the norm.

This yields a practical rule for evaluating Ukrainian. Present each item in both orders, report the agreement between them, and read the prompted accuracy only for models that clear a stated consistency level; for the rest, probability scoring is the channel that measures grammar. The extra cost is one additional presentation per item.

## 4.3. Knowing Versus Using

The comparison between the two scoring methods is meaningful only where the prompted channel responds to content. We fixed a threshold of 0.6 order consistency in advance; four models pass it: MamayLM (0.759), google/gemma-3-4b-it (0.666), Qwen2.5-7B-Instruct (0.657) and Llama-3.1-8B-Instruct (0.628). For reference, a model that picked between the two letters at random would reach a consistency of about 0.5, so values near zero reflect a systematic preference for one answer position. The threshold of 0.6 is a convention, and the set of models it selects changes little around it: at 0.55 it adds lapa-v0.1.3-instruct, discussed below as the borderline case, and at 0.7 it keeps MamayLM alone, whose two channels agree ($p = 1.00$).

Among these four, McNemar's test on the paired outcomes finds no difference between the methods for MamayLM ($p = 1.00$), Llama-3.1-8B-Instruct ($p = 0.90$) or Qwen2.5-7B-Instruct ($p = 0.07$), but does find one for gemma-3-4b-it ($p = 0.002$), which is more often right by probability than by prompt: it uses the norm without stating it on 10.4% of pairs, and states it without using it on 6.9%. So the dissociation reported for English by Hu and Levy [9] is present here too, but it is neither universal nor large: three of four interpretable models show agreement between the channels, and the fourth shows a modest gap in the direction their work predicts.

One borderline case deserves mention. lapa-v0.1.3-instruct has consistency 0.572, just below the threshold, and is therefore excluded from this comparison by the rule fixed beforehand. Its numbers point the other way — prompted accuracy 0.986 against probability accuracy 0.886, and 0.994 on its order-consistent subset.

### 4.4. Ukrainian Adaptation, Measured Against the Model's Own Starting Point

Two of the adapted models can be compared directly with the checkpoint they were trained from, on the same pairs and under identical loading; Table 4 reports both comparisons.

**Table 4**
Ukrainian adaptation against the exact checkpoint it started from: probability accuracy on the same pairs, under the same loading configuration. The paired tests are reported in the text.

| Phenomenon | Base | Adapted | Delta |
|---|---|---|---|
| **gemma-3-4b-it → MamayLM-Gemma-3-4B-IT** | | | |
| Subject–verb (control) | 0.997 | 1.000 | +0.003 |
| Adjective agreement | 0.897 | 0.987 | +0.090 |
| Numeral + noun | 0.907 | 0.957 | +0.050 |
| Verb government | 0.833 | 0.733 | -0.100 |
| Calques | 0.772 | 0.940 | +0.168 |
| Vocative | 0.247 | 0.510 | +0.263 |
| All phenomena | 0.766 | 0.868 | +0.102 |
| **gemma-3-12b-pt → lapa-12b-pt** | | | |
| Subject–verb (control) | 1.000 | 1.000 | +0.000 |
| Adjective agreement | 0.970 | 0.993 | +0.023 |
| Numeral + noun | 0.937 | 0.983 | +0.047 |
| Verb government | 0.783 | 0.783 | +0.000 |
| Calques | 0.853 | 0.951 | +0.098 |
| Vocative | 0.267 | 0.560 | +0.293 |
| All phenomena | 0.801 | 0.889 | +0.088 |

lapa-12b-pt is a continued pre-train of google/gemma-3-12b-pt, which makes it the cleaner of the two comparisons. Overall probability accuracy rises from 0.801 to **0.889**. McNemar on the paired outcomes: 164 pairs corrected by adaptation against 37 broken, chi2 = 79.0, p = 6.3e-19. The instruction-tuned lapa-v0.1.3-instruct gives essentially the same picture against the same base (0.886, chi2 = 77.1, p = 1.6e-18).

MamayLM against google/gemma-3-4b-it replicates it at a smaller scale: 0.766 to **0.868**, with 178 pairs corrected against 31 broken, chi2 = 102.0, p = 5.6e-24. One caveat applies to this pair: the MamayLM model card lists both gemma-3-4b-pt and gemma-3-4b-it as base models, since the adaptation used data mixing and model merging, so the starting point is not a single checkpoint. The Lapa comparison does not have this complication.

The structure of the gain is the same in both pairs, and it is the substantive finding. The control phenomenon does not move: 1.000 to 1.000 for Lapa, 0.997 to 1.000 for MamayLM. The gain is concentrated where multilingual models fail systematically — the vocative (+0.293 and +0.263) and calques (+0.098 and +0.168). Ukrainian adaptation does not raise everything a little; it repairs specific deficits.

Verb government behaves differently in the two pairs: it falls by 0.100 for MamayLM and does not move for Lapa (0.783 in both). On 60 pairs a change of that size lies within the range chance produces, and no comparable effect appears in the second pair, so we treat it as a property of one model and not as a cost of adaptation.

## 4.5. Secondary Observations

**Overt and covert calques do not differ.** We expected covert calques — fluent, structurally Ukrainian constructions borrowed semantically, such as «порушити питання» — to be harder for models than overt russianisms such as «гаразд». They are not: 0.778 against 0.761 under probability scoring. The hypothesis is not supported at the aggregate level. Section 4.7 examines the items individually and shows that the aggregate hides a different and sharper split, which the overt/covert distinction does not capture.

**Instruction tuning does not improve probability accuracy.** Averaged over the multilingual models, base checkpoints reach 0.702 and instruction-tuned ones 0.672. Of six base/instruct pairs of the same checkpoint, five decline after tuning, most sharply Llama-3.2-1B (0.684 to 0.578) and Llama-3.2-3B (0.717 to 0.640); only Qwen2.5-3B improves, and marginally (0.628 to 0.640). Combined with the consistency results in 4.2, the picture is that instruction tuning improves a model's ability to answer questions about the norm without improving its internal representation of the norm.

**Size predicts little.** gemma-2-2b (0.740) outperforms Qwen2.5-7B-Instruct (0.651) despite being smaller by a factor of three. Within the Gemma family scale does help — 1B 0.717, 2B 0.740, 4B 0.766, 12B 0.801 — but across families the composition of the training data dominates. The three Ukrainian-adapted models occupy the top three places overall (0.889, 0.886, 0.868), ahead of every multilingual model including the 12B one.

## 4.6. The Full Model-by-Phenomenon Matrix

Averages hide structure, and three patterns are visible only in the complete matrix of Table 5.

**Table 5**
Probability accuracy for every model on every phenomenon, ordered by overall accuracy. (ua) marks Ukrainian-adapted models. Ctrl = subject–verb agreement (control), Adj = adjective agreement, Num = numeral–noun agreement, Gov = verb government, Calq = calques, Voc = vocative.

| Model | Ctrl | Adj | Num | Gov | Calq | Voc | All |
|---|---|---|---|---|---|---|---|
| lapa-12b-pt (ua) | 1.00 | .99 | .98 | .78 | .95 | .56 | .89 |
| lapa-v0.1.3-instruct (ua) | 1.00 | .99 | .99 | .80 | .89 | .58 | .89 |
| MamayLM-Gemma-3-4B-IT (ua) | 1.00 | .99 | .96 | .73 | .94 | .51 | .87 |
| gemma-3-12b-pt | 1.00 | .97 | .94 | .78 | .85 | .27 | .80 |
| gemma-3-4b-it | 1.00 | .90 | .91 | .83 | .77 | .25 | .77 |
| gemma-2-2b | 1.00 | .93 | .90 | .87 | .84 | .05 | .74 |
| Llama-3.1-8B-Instruct | .97 | .88 | .94 | .82 | .76 | .13 | .74 |
| gemma-2-2b-it | .97 | .87 | .88 | .77 | .83 | .09 | .72 |
| Llama-3.2-3B | .97 | .89 | .91 | .85 | .78 | .03 | .72 |
| gemma-3-1b-pt | .97 | .91 | .86 | .82 | .84 | .03 | .72 |
| gemma-3-1b-it | .97 | .82 | .79 | .78 | .84 | .11 | .70 |
| Llama-3.2-1B | .92 | .85 | .88 | .85 | .76 | .01 | .68 |
| Qwen2.5-7B-Instruct | .98 | .84 | .74 | .65 | .68 | .02 | .65 |
| Llama-3.2-3B-Instruct | .85 | .82 | .76 | .80 | .70 | .06 | .64 |
| Qwen2.5-3B-Instruct | .89 | .82 | .78 | .73 | .68 | .02 | .64 |
| Qwen2.5-1.5B | .98 | .74 | .78 | .63 | .62 | .02 | .63 |
| Qwen2.5-3B | .90 | .82 | .75 | .68 | .66 | .02 | .63 |
| Qwen2.5-1.5B-Instruct | .92 | .76 | .74 | .65 | .62 | .02 | .61 |
| Llama-3.2-1B-Instruct | .79 | .72 | .59 | .75 | .66 | .12 | .58 |

**The vocative is the only systematic failure.** Across 19 models and six phenomena there is not a single cell below 0.5 other than in the vocative column. Every model is above chance on agreement, on numerals, on government and on calques; the weakest non-vocative cell in the entire table is 0.593 (Llama-3.2-1B-Instruct on numeral–noun agreement). The vocative column, by contrast, runs from 0.013 to 0.577 — and its ceiling is reached by a Ukrainian-adapted model. No model in the study reaches 0.6 on the vocative.

**Verb government is the one phenomenon Ukrainian adaptation does not repair.** It is also the flattest column, ranging only from 0.633 to 0.867, and its best score belongs to gemma-2-2b — a 2B multilingual model that outperforms all three Ukrainian-adapted models on it (0.733 to 0.800). This is consistent with the controlled pairs, where adaptation changed government by −0.100 in one case and 0.000 in the other. Whatever the adaptation data contains, it is not improving case selection after verbs, and the phenomenon appears to depend on something other than exposure to Ukrainian text volume.

**Instruction tuning degrades even the control.** Three models fall below 0.9 on subject–verb agreement, and all three are instruction-tuned: Llama-3.2-1B-Instruct (0.790), Llama-3.2-3B-Instruct (0.853) and Qwen2.5-3B-Instruct (0.893). In each case the corresponding base checkpoint scores higher — 0.917, 0.973 and 0.900 respectively. Since the control is the phenomenon we use to confirm that the harness works, this also qualifies that confirmation: for these three models the sanity check is weaker, and their other scores should be read with that in mind.

## 4.7. Which Calques, and Why

The aggregate calque score of 0.773 describes a mixture, not a level, and Table 6 lists the items that drive the failures. Across the 46 curated items the distribution is strongly bimodal: the median item scores 0.888, five items are answered correctly by every model, and only seven fall below 0.5. Models are not uniformly mediocre on calques; they solve most of them and fail badly on a small, specific tail.

**Table 6**
Calque items below 0.5 probability accuracy, with the non-normative variant that models prefer, scored separately for the multilingual and the Ukrainian-adapted models. 7 of the 46 items are below 0.5.

| Normative | Type | Example of the rejected form | Multiling. | Ukr. |
|---|---|---|---|---|
| порушити питання | subtle | Депутат підняв питання про бюджет. | 0.16 | 0.25 |
| гаразд | overt | Ладно, я згоден. | 0.09 | 0.75 |
| бувай | overt | Пока, до зустрічі. | 0.25 | 0.33 |
| перепрошую | subtle | Вибачаюсь, я запізнився. | 0.22 | 0.67 |
| брати до уваги | subtle | Прошу прийняти це до уваги. | 0.31 | 0.58 |
| наводити приклад | subtle | Викладач приводить приклад. | 0.36 | 1.00 |
| складати іспит | subtle | Завтра я здаю іспит. | 0.41 | 0.92 |

That tail is not the one our design predicted. Section 4.5 reported that overt russianisms and covert calques do not differ in aggregate, and the item-level view explains why the distinction was the wrong axis: of the seven hardest items, five are covert and two overt, and they are separated from the easy ones by something else entirely. What the hard items share is that the form the models prefer is the form Ukrainian speakers actually use — «підняти питання» instead of «порушити питання», «ладно» for «гаразд», «пока» for «бувай», «вибачаюсь» for «перепрошую», «здавати іспит» for «складати іспит». The difficulty tracks the corpus frequency of the non-normative variant, not its structural transparency. A model trained on web text has seen «підняти питання» far more often than the normative alternative, and its probabilities reflect that faithfully.

Ukrainian adaptation repairs most of this tail, and the exceptions are informative. Two items are essentially fixed — «наводити приклад» rises from 0.36 in the multilingual models to 1.00 in the Ukrainian-adapted ones, and «складати іспит» from 0.41 to 0.92 — and «гаразд» improves from 0.09 to 0.75. But «порушити питання» moves only from 0.16 to 0.25, and «бувай» from 0.25 to 0.33. The two that resist are the two whose non-normative variants dominate written Ukrainian as well as spoken: «підняти питання» is standard in journalistic prose, and «пока» is ubiquitous in informal writing. Where the non-normative form is merely colloquial, additional Ukrainian pre-training displaces it; where it has colonised the written register, it does not.

This has a direct practical consequence. For a deployed system, the calque problem is not diffuse but concentrated in a handful of entrenched items, and a lexical post-check over a list of that size is cheap. Model priors alone should not be relied on for them, in multilingual and Ukrainian-adapted models alike.

# 5. Limitations

## 5.1. What the Pairs Are and Are Not

The pairs are generated from templates, not drawn from corpora. This is what makes the comparison controlled — the two members of a pair differ in one edited span and in nothing else — but it also means the sentences are short, syntactically simple and stylistically uniform. A model that handles the vocative correctly in «Олеже, зайди» might behave differently in a long sentence with competing attractors, and our results do not speak to that. The frames themselves also carry frequency effects we did not control: eight address frames recur across all 300 vocative pairs, so a model's familiarity with those particular frames contributes to its score.

The benchmark measures agreement with a codified standard, not acceptability to speakers. Where reference works disagree, or where an alternation marks nuance instead of norm, items were excluded during construction. That filtering is not perfect, and borderline items may remain. A benchmark of normativity is necessarily narrower than the language it is drawn from.

Two phenomena rest on small samples. Verb government has 60 pairs and calques 184; per-model estimates on the former carry wide intervals, which is why we declined to interpret the 0.100 drop observed for one adapted model.

## 5.2. Scope of the Methodological Claim

Our finding about the prompted channel is specific to the format we tested: two sentences, labelled with Cyrillic letters, one question template in Ukrainian, answers scored by comparing the log-probabilities of the two letters. Few-shot prompting, chain-of-thought, free-form generation with extraction, English instructions, and multiple-choice with more than two options may all behave differently. Galeshchuk et al. report that few-shot examples help pretrained models follow the answer format, which is precisely the mechanism that would raise order consistency; we did not test it.

The protocol we propose is correspondingly scoped: within this format, order consistency establishes which models support a prompted evaluation and which require probability scoring. Whether the same diagnostic is needed for other prompt formats is an empirical question our design can answer, and applying it to few-shot and generative formats is the natural next step.

## 5.3. Numerical Precision and Hardware

Models were run under different numerical configurations dictated by a 16 GB accelerator: models up to 3B in their native precision, larger models in 4-bit NF4 quantisation, and the 12B checkpoints in float32 because Gemma 3 activations overflow float16 at that scale. Within each controlled base–adapted pair the configuration is identical, so those comparisons are unaffected. Comparisons across families are not equally protected: we did not measure how much 4-bit quantisation perturbs sentence log-probabilities, and a difference of a few points between two models loaded differently should not be read as a difference in competence.

The float16 overflow deserves a note of its own, since it produced silent failure instead of an error. All 1,444 log-probabilities returned NaN for three 12B models, which a pipeline that simply recorded scores would have entered into the results as uniformly incorrect predictions. Evaluation harnesses running Gemma 3 on hardware without bfloat16 support should check for this explicitly.

### 5.4. Coverage of Models

Only open-weight models are evaluated, because probability scoring requires access to token log-probabilities. The models most widely deployed in Ukrainian-language products are often closed and reachable only through an interface, and for those the prompted channel — the one we show to be unreliable at small scale — is the only one available. Our results therefore describe what can be measured, not necessarily what is deployed. Whether the vocative inversion holds for large closed models is an open question that this instrument cannot answer.

Each model was evaluated once. Probability scoring is deterministic given a fixed configuration, so this introduces no sampling variance, but it also means we cannot report variability due to quantisation seeds or kernel-level non-determinism.

## 6. Conclusion

We released 1,444 Ukrainian minimal pairs across six phenomena of normative usage, validated by two native speakers, and used them to evaluate 19 language models under two scoring methods.

The protocol is the main contribution. Probability scoring over minimal pairs measures normative usage for any open-weight model, and the order-consistency diagnostic establishes when a prompted score can be read as grammar. For every base checkpoint we tested, and for instruction-tuned models below roughly 4B, the prompted number is governed by answer position; ten of nineteen models answered by content on fewer than 5% of pairs, a 12B model among them. One extra presentation per item is enough to detect this, and we recommend reporting it wherever prompted or multiple-choice evaluation of Ukrainian is used.

The central empirical finding is the vocative inversion. Multilingual models do not merely fail on the vocative; they prefer the non-normative form, at 0.079 accuracy against 0.951 on the control. This is a concrete and actionable defect for any system that addresses users by name.

Ukrainian adaptation substantially reduces it. Measured against their own base checkpoints, lapa-12b-pt gains 0.088 overall and MamayLM 0.102, with the improvement concentrated in the vocative and calques while the control stays at ceiling. For MamayLM, whose adaptation also involved model merging, the gain cannot be attributed to Ukrainian data alone.

One qualification bounds the remedy, and it follows from the item-level results. The adaptation corpora of both Ukrainian models are large Ukrainian web collections, not curated normative text, and what they displace is what is rare in multilingual training data, not what is wrong. Where the non-normative variant is itself frequent in Ukrainian writing, additional Ukrainian pre-training leaves it in place: «підняти питання» is preferred over «порушити питання» by the multilingual models at 0.16 and by the Ukrainian-adapted ones at 0.25. More Ukrainian data displaces what is rare, not what is entrenched, and the entrenched residue is small enough to be handled by a lexical check.

Three directions follow. The pair set covers six phenomena and could be extended to participles, aspect and word order. The prompted channel was tested with a single question template, and other formats may behave differently. And the finding that order consistency follows instruction tuning and not scale deserves study in its own right, across a wider range of models and tuning recipes than we could evaluate here.

## Declaration on Generative AI

During the preparation of this work, the authors used Claude (Anthropic) in order to: Generate literature review, Paraphrase and reword, Formatting assistance. Further, the authors used Claude (Anthropic) for figures 1, 2 and 3 in order to: write the code that produces them from the recorded scores. The study design, the curated word lists, the human validation of the pairs and all scientific claims are the authors' own. After using this tool, the authors reviewed and edited the content as needed and take full responsibility for the publication's content.